\documentclass[letterpaper, 10 pt, conference]{ieeeconf}  % Comment this line out if you need a4paper

\IEEEoverridecommandlockouts                              % This command is only needed if 
\usepackage{cite}
\usepackage{bm}
\usepackage{hyperref}
\hypersetup{
    colorlinks=true,
    linkcolor=black,
    filecolor=black,      
    urlcolor=blue,  % ここでURLの色を設定できます
    citecolor=black,
}
\usepackage{epic,eepic}
\usepackage{url}
\usepackage{lipsum}
\usepackage{pifont}
\usepackage{amsmath,amssymb}
\usepackage{amssymb}
\usepackage{txfonts}
\usepackage{nccmath}
\usepackage{comment}
\usepackage{balance}
\usepackage{multirow}
\usepackage{textcomp}
\usepackage{multirow}
\usepackage{subcaption}
\usepackage{booktabs}
\usepackage{siunitx}
\usepackage{tabularx}
\usepackage{threeparttable}
\usepackage{algorithm}
\usepackage{dblfloatfix}
\usepackage{CJKutf8}
\usepackage[table]{xcolor}
\usepackage[noend]{algpseudocode}
\usepackage{xcolor}
\usepackage{eso-pic} % Acceptance notice on the first page

\newcommand{\EQ}[1]{Eq.~(\ref{#1})}
\usepackage{graphicx}
\title{\LARGE \bf
KING: Embodiment-Aware Kinematic Graph Neural Network for Unified Motion Representation of Legged and Wheeled Robots
}

\author{Taku Okawara$^{1}$, Aoki Takanose$^{1}$, Kenji Koide$^{1}$, Shuji Oishi$^{1}$, and Masashi Yokozuka$^{1}$ % <-this % stops a space
\thanks{*This work is supported in part by JST K Program JPMJKP23G2, the BRIDGE Program (R7-H05), JSPS KAKENHI Grant Number JP26K21361, and the Suzuki Foundation.}% <-this % stops a space
\thanks{$^{1}$All the authors are with the Department of Information Technology and
Human Factors, the National Institute of Advanced Industrial Science and
Technology, Tsukuba, Ibaraki, Japan, taku.okawara@aist.go.jp}%
}

\begin{document}

\maketitle
\thispagestyle{empty}
\pagestyle{empty}

% arXiv acceptance notice, positioned as in arXiv:2505.12537v2.
% The starred command limits the notice to the first page.
\AddToShipoutPictureBG*{
  \AtPageUpperLeft{%
    \put(0,-40){\raisebox{15pt}{\makebox[\paperwidth]{\begin{minipage}{21cm}\centering
      \textcolor{gray}{This article has been accepted for publication in the proceedings of the \\
      2026 IEEE/RSJ International Conference on Intelligent Robots and Systems (IROS)}
    \end{minipage}}}}%
  }
}

%%%%%%%%%%%%%%%%%%%%%%%%%%%%%%%%%%%%%%%%%%%%%%%%%%%%%%%%%%%%%%%%%%%%%%%%%%%%%%%%
\begin{abstract}
Kinematic models provide reliable motion constraints for odometry estimation in featureless environments, where exteroceptive sensing degrades and IMU integration drifts. Learning-based kinematic models can achieve more accurate odometry estimation than model-based methods by capturing nonlinear effects; however, most existing learning-based models are trained on a single embodiment and generalize poorly to new embodiments. This generalization is difficult because the meanings and structures of proprioceptive measurements vary across embodiments, including the number of joints and ground-contact elements (e.g., wheels, feet). To address this challenge, we propose KING, a Graph Neural Network (GNN)-based kinematic model that explicitly incorporates robot embodiments by representing them as a common graph. We show that wheel and leg kinematic models can be expressed by a unified representation, enabling a single model for both wheeled and legged robots.
Trained on datasets spanning diverse embodiments, KING provides a unified representation of wheeled and legged kinematics and achieves high-accuracy odometry estimation in real environments. KING estimates accurate odometry using only an embodiment description (e.g., a URDF file) and on-board proprioception (encoders and an IMU) and can be adapted to new robot embodiments through few-shot learning with only one minute of data, avoiding retraining from scratch on a new dataset for each robot.
The project page is available at: \url{https://smrg-aist.github.io/king_project_page/}
\end{abstract}

%%%%%%%%%%%%%%%%%%%%%%%%%%%%%%%%%%%%%%%%%%%%%%%%%%%%%%%%%%%%%%%%%%%%%%%%%%%%%%%%
\section{INTRODUCTION}

Robust odometry estimation is essential for reliable autonomous navigation systems.
To enhance its robustness under challenging conditions, it is important to fuse complementary constraints from exteroceptive sensors (e.g., LiDAR and cameras), IMUs, and kinematic models.
Exteroceptive sensor-based constraints become unreliable in featureless environments because feature matching of these measurements fails.
IMU-based constraints are also unreliable under long-term featureless conditions due to integration of noisy measurements (particularly double integration of linear accelerations).
In contrast, kinematic model-based constraints can provide reliable motion constraints from encoder measurements (e.g., joint angles and angular velocities) with less integration effort than IMU measurements.
However, typical kinematic models can degrade in the presence of terrain-dependent phenomena that are difficult to model rigorously, such as wheel/foot slip and terrain deformation~\cite{wisth2022vilens,okawara2024neuralwheel,okawara2025tightlyleg}.

\begin{figure}[tb]
  \centering
  \includegraphics[width=1.00\linewidth]{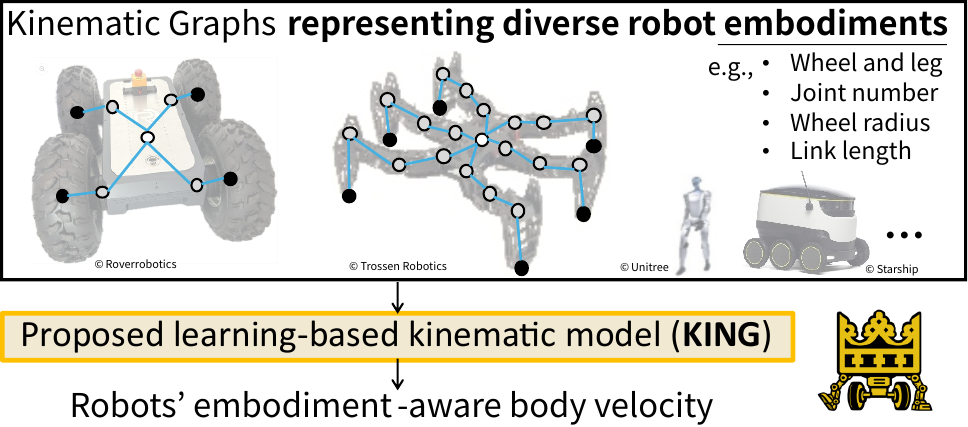}
  \caption{Robot embodiment-aware learning-based kinematic model (KING), which is applied to various mobile robots.}
  \label{fig:fig1}
\end{figure}

Learning-based kinematic models have been shown to better capture such nonlinear effects by implicitly modeling complex robot--terrain interactions~\cite{wassermanlegolas,onyekpe2021whonet,van2022learning}.
However, their generalization is often limited to specific terrains and a single robot \emph{embodiment}\footnote{In this paper, \emph{morphology} denotes the robot topology (e.g., a quadruped or a 4WD skid-steering configuration), whereas \emph{embodiment} denotes a particular instance of a morphology together with its continuous kinematic parameters (e.g., wheel radii, link lengths, and IMU placement).}.
Achieving a kinematic model with generalization requires designs of model architecture and input/output representations that explicitly encode both embodiment and terrain context, which in turn demand diverse training data.
For terrain generalization, Okawara \textit{et al.} pretrained an MLP to represent terrain-dependent features from data collected across diverse surfaces, and then adapted the terrain module online to the current ground conditions~\cite{okawara2024neuralwheel,okawara2025tightlyleg}.
In contrast, robot embodiment generalization is challenging because the \emph{inputs} to a learned kinematic model---low-level information such as proprioceptive measurements and embodiment parameters---vary across embodiments in meaning and dimensionality with the numbers of joints and contact elements (e.g., wheels and feet).
Accordingly, prior cross-embodiment learning studies~\cite{xiao2025anycar,o2024open,rath2025xmop} typically rely on higher-level abstractions (e.g., body velocity estimated by SLAM, end-effector motion in its local frame) instead of directly sharing low-level motion representations across heterogeneous morphologies.

We tackle embodiment generalization of learning-based kinematic models for legged and wheeled robots in this work.
To address this challenge, we first show that kinematic models of wheeled and legged robots can be expressed in a unified formulation using manipulators' kinematic model.
Thanks to this interpretation, we encode diverse robots' embodiments and local proprioceptive motion signals (e.g., encoder and IMU measurements) as a common graph, where nodes contain information about embodiment parameters and proprioceptive measurements, and edges connect nodes according to each robot's embodiment-specific kinematic connectivity.
By training a graph neural network (GNN) on these variable-structure graphs, we obtain a single kinematic model, termed \textit{KING: embodiment-aware KINematic Graph neural network}, that can be deployed across diverse wheeled and legged robots.
This formulation enables KING to overcome the above challenges by encoding embodiment-dependent low-level inputs into a shared graph representation, thereby learning a single learning-based kinematic model that generalizes across diverse robot embodiments.

Our contributions are summarized as follows:
\begin{enumerate}
  \item We derived a unified kinematic model formulation for both wheeled and legged robots.
        This formulation provides a principled basis for describing them within the common kinematic framework, enabling a single learned model to cover these distinct morphologies.
  \item We introduced this unified formulation into a common graph of kinematic structures and proprioceptive information across robot morphologies.
        Using this graph as the input to a GNN, we proposed KING, an embodiment-aware kinematic model applicable to both legged and wheeled robots.
  \item We demonstrated KING's generalization to unseen robot morphologies through cross-validation and showed that few-shot sim-to-real adaptation enables accurate odometry estimation on real robots.
\end{enumerate}

\section{Related works}

\subsection{Learning-based kinematic models}
\label{subsection:lbkm}
\textbf{Kinematic model for specific conditions: }
Learning-based kinematic models~\cite{wassermanlegolas,onyekpe2021whonet,van2022learning} have been widely studied to well capture terrain-dependent effects that are difficult to model explicitly, such as slip and ground deformation.
These models take proprioceptive sensor measurements (e.g., encoder and IMU measurements) as inputs and estimate the robot body's motion.
These works demonstrated that learning-based kinematic models outperformed model-based methods in odometry estimation accuracy because nonlinear robot-terrain interactions are implicitly represented.

These approaches typically adopt fixed-size architectures (e.g., MLPs, RNNs, or CNNs).
Because their input dimensionality depends on the robot's morphology and the number of DOF, redesigning the input representation is required when transferring such models across embodiments.
Consequently, when a pretrained model is applied to a different platform or terrain---or even to the same platform after changes to embodiment parameters---additional data collection and retraining are often required.
As a result, many learning-based kinematic models are specialized to specific conditions (i.e., a single-robot embodiment and terrain), and their generalizability across conditions remains limited.

\textbf{Kinematic models for enhanced generalization: }
% Several works aim to improve the generalization of learning-based kinematic models.
Several works aim to improve the generalization of learning-based kinematic models across terrains and robot embodiments by collecting diverse datasets and designing models that reflect terrain and robot-embodiment information.
Okawara~\cite{okawara2024neuralwheel,okawara2025tightlyleg} addressed terrain generalization by collecting various terrain types of datasets and introducing a terrain-dependent MLP and adapting it online to the current ground condition.
This online adaptation mitigates the need for data collection and batch retraining when a robot moves onto new terrain; however, these methods still assume a single-robot embodiment.

For robots' embodiment generalization, AnyCar~\cite{xiao2025anycar} learns a dynamics model for robust localization and motion prediction across Ackermann-steered vehicles with varying body sizes and wheel radii.
However, AnyCar relies on exteroceptive sensor-based state estimation (e.g., SLAM) to represent vehicle motions across different sizes; namely, the embodiment information (e.g., wheel radii, body sizes) is not explicitly incorporated in the model.
Thus, this method cannot be extended to other wheel configurations (e.g., six-wheeled skid-steering robots) or to legged robots.

\subsection{Cross-Embodiment Learning}
\label{subsection:crossembodiment}
While most cross-embodiment learning studies focus on policy model learning, our method learns a kinematic model that generalizes across diverse mobile robot embodiments; nonetheless, cross-embodiment learning techniques for embodiment generalization are closely related to our work.
One of the most important key operations of cross-embodiment learning is to unify data representations across different morphologies, so that observations and actions preserve consistent meanings despite variations in embodiment parameters and joint DOFs.
To describe motions of diverse robots with a single model, many approaches rely on task-space quantities~\cite{o2024open,rath2025xmop} or exteroceptive modalities~\cite{shah2022gnm,xiao2025anycar}, which are weakly dependent on the robot embodiment and can be shared across embodiments.
General Navigation Model~\cite{shah2022gnm} demonstrates zero-shot transfer of vision-based navigation policies across various mobile robots by using multiple camera images as inputs.
RT-X~\cite{o2024open} standardizes action representations across diverse robots by expressing actions in task-space frames, such as end-effector-frame motions for manipulators and base-frame motions for mobile robots.
However, because these methods do not explicitly encode robot embodiment, they often omit embodiment parameters, joint-space states, and their constraints, which are important for robot-specific control.

In contrast to the aforementioned related works, LocoFormer~\cite{liu2025locoformer} processes joint-space states via \textit{Unified observation}, which concatenates joint states from diverse robots into a fixed-length vector.
While the unified observation enables training a shared policy across multiple robot morphologies, it does not explicitly encode embodiment structure (e.g., kinematic connectivity) and relies on a hand-designed joint superset with fixed size, which may limit generalization to different morphologies.
In contrast, NerveNet~\cite{wang2018nervenet} and GCNT~\cite{ijcai2025p972} explicitly represent a robot's morphology as a graph, where nodes correspond to body components (e.g., joints/links) and edges reflect kinematic connectivity.
By training a GNN on such morphological graphs, these methods can train their models conditioned on the robot's morphology.
While these methods learn morphology-conditioned policies across diverse legged robots, they do not explicitly study a unified formulation that spans both wheeled and legged robots and fully incorporates low-level motion data.

Learning kinematic models that strongly depend on embodiment is inherently challenging to generalize across robots, because both the meanings and dimensionality of proprioceptive signals and physical parameters vary with morphology.
In contrast to prior cross-embodiment learning works that avoid such low-level, embodiment-dependent information, our method explicitly represents robot embodiments using a graph grounded in the Unified Kinematic-Model Representation.
The proposed representation enables a single GNN to learn a kinematic model across diverse embodiments while preserving the meaning of proprioceptive measurements and embodiment parameters.
As a result, our approach supports cross-embodiment learning using low-level information and provides a unified model that spans substantially different robot morphologies, including both legged and wheeled robots.

\begin{figure}[tb]
  \centering
  \includegraphics[width=1.00\linewidth]{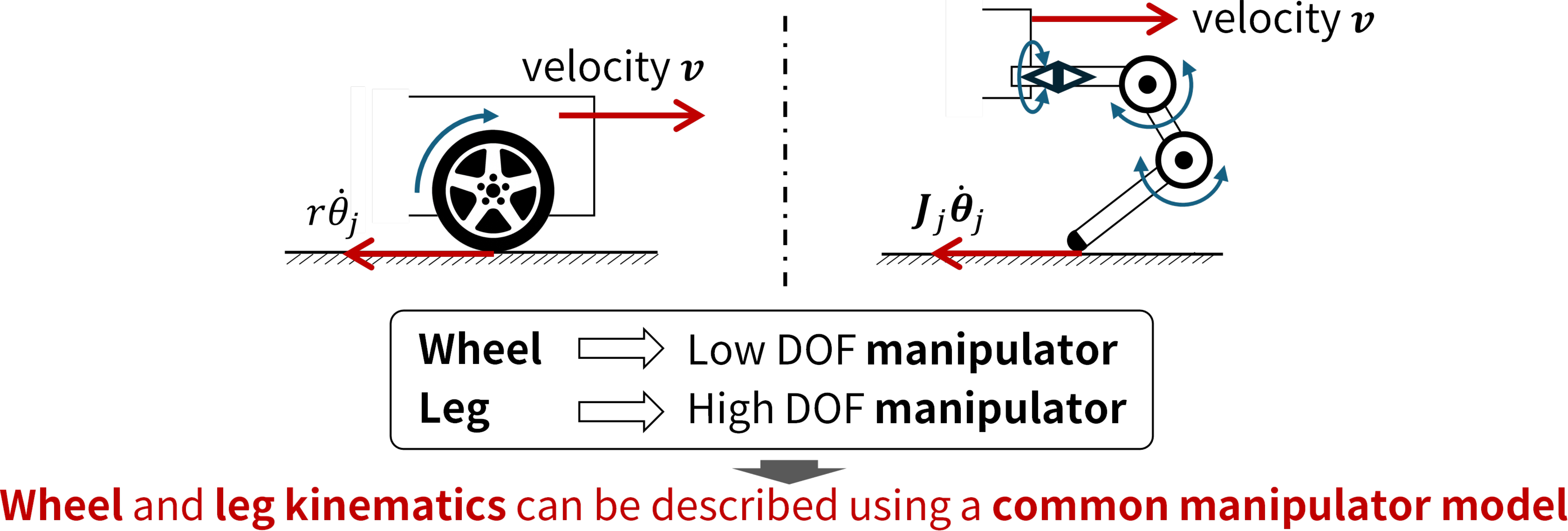}
  \caption{Kinematic model similarity for wheels and legs.}
  \label{fig:wheel_and_leg_diagram}
\end{figure}

\section{A Unified Kinematic-Model Representation for Legged and Wheeled Robots}\label{section:UKMR}
In this section, we show that kinematic models of wheeled and legged robots can be expressed within a single common formulation. 
This derivation provides a theoretical basis for treating both classes of robots within a unified kinematic model representation, which we later exploit to learn a single model across diverse morphologies.

\subsection{Kinematic model overviews of legged and wheeled robots}
We first summarize standard kinematic models for legged and wheeled robots.

\textbf{Legged robots:}
The kinematics of a leg can be described in the same manner as a serial robot manipulator.
Specifically, an $N$-legged robot's body translational velocity ${}^{\mathrm{L}}\!\bm{v}_{\mathrm{B}}$ is described by a $j$-th foot (i.e., end-effector) velocity $\bm{J}_j \dot{\bm{\theta}}_j$, described as Eq.~\ref{eq:leg_odom}~\cite{wisth2022vilens,yang2023cerberus}:
\begin{align}
  {}^{\mathrm{L}}\!\bm{v}_{\mathrm{B}} = - \bm{J}_j \, \dot{\bm{\theta}}_j,
  \label{eq:leg_odom}
\end{align}
where $\bm{J}_j$ is a Jacobian matrix related to $j$-th foot position and joint angles, $\dot{\bm{\theta}}_j$ is joint angular velocities of $j$-th leg.
Note that Eq.~\ref{eq:leg_odom} is satisfied in the assumptions, where the foot does not slip, and the ground is not deformed.

\textbf{Wheeled robots:}
In the case of an $N$-wheeled robot, its body translational velocity ${}^{\mathrm{W}}\!{v}_{\mathrm{B}}$ is described by a $j$-th wheel velocity $r \dot{\theta}_j$, described as Eq.~\ref{eq:wheel_odom}:
\begin{align}
  {}^{\mathrm{W}}\!{v}_{\mathrm{B}} = - r \dot{{\theta}}_j,
  \label{eq:wheel_odom}
\end{align}
where $r$ is the wheel radius, $\dot{\theta}_j$ is the wheel angular velocity for the $j$-th wheel.
%, $m$ is the number of degrees of freedom (DOF) of its leg.

In both legged and wheeled systems, the final body velocity is typically estimated as the mean velocity of all feet/wheels contacting the ground.

\subsection{A unified kinematic representation based on a kinematic model of manipulators} \label{subsec:unified_kinematic_representation}
Equations~\eqref{eq:leg_odom} and~\eqref{eq:wheel_odom} share a common structure: the body velocity can be expressed in terms of the velocity at a ground-contact point (a foot or wheel contact), which is formulated as end-effector velocity in manipulator models.
This suggests a unified interpretation in which a leg is a higher-DOF ``manipulator'' whose end-effector periodically contacts the ground, whereas a wheel is a lower-DOF ``manipulator'' whose contact is (ideally) continuous.
A key practical distinction between legs' and wheels' motion lies in their contact patterns: legs typically exhibit periodic ground contact due to high DOFs, whereas wheels maintain continuous contact with the ground due to low DOFs.

As shown in Eq.~\ref{eq:leg_odom}, the foot velocity is described by a linear combination of the Jacobian matrix elements (e.g., joint angles, link length) and multiple joint angular velocities $\dot{\theta}_{j,1}, \dot{\theta}_{j,2},..., \dot{\theta}_{j,m}$.
Since $\bm{J}_j \dot{\bm{\theta}}_j$ is a linear combination, eliminating joints removes the corresponding terms and results in a lower-DOF case.
Eq.~\eqref{eq:wheel_odom} can be viewed as the 1-DOF special case, where the Jacobian matrix and joint angular velocity reduce to a single scalar term $r$ and $\dot{\theta}_j$, respectively.
Therefore, both leg and wheel kinematic models can be represented within a unified formulation using manipulators' kinematics.
This unified representation lets the proposed GNN model encode both legged and wheeled kinematics in a shared input space, so their motion data can contribute to learning the same underlying structure rather than being treated as fundamentally mismatched.

\begin{figure}[tb]
  \centering
  \includegraphics[width=0.85\linewidth]{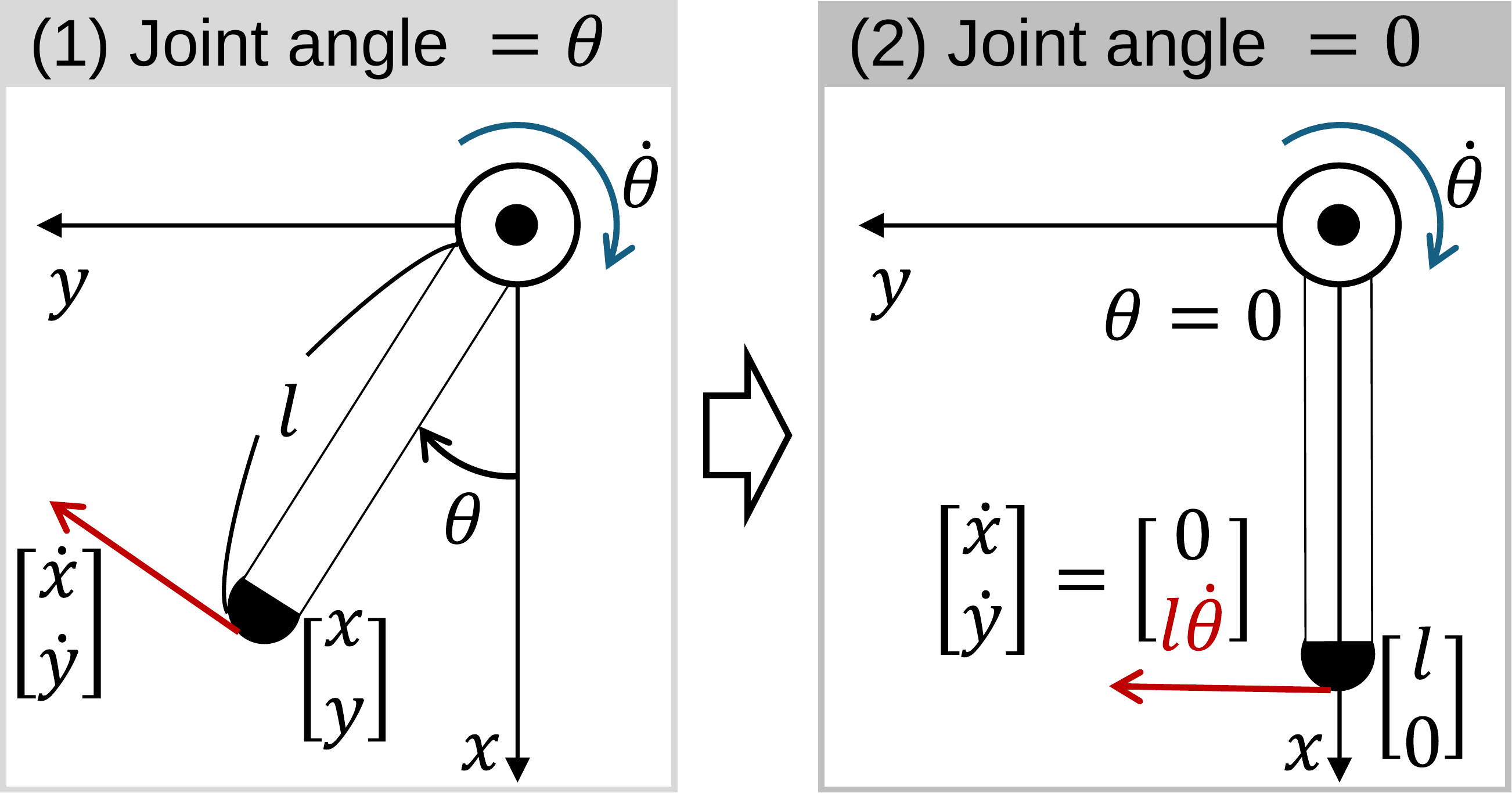}
  \caption{Certifying that the motion of a wheel is equivalent to that of a 1~DOF manipulator.}
  \label{fig:1DOF_arm}
\end{figure}

\begin{figure*}[tb]
  \centering
  \includegraphics[width=1.0\linewidth]{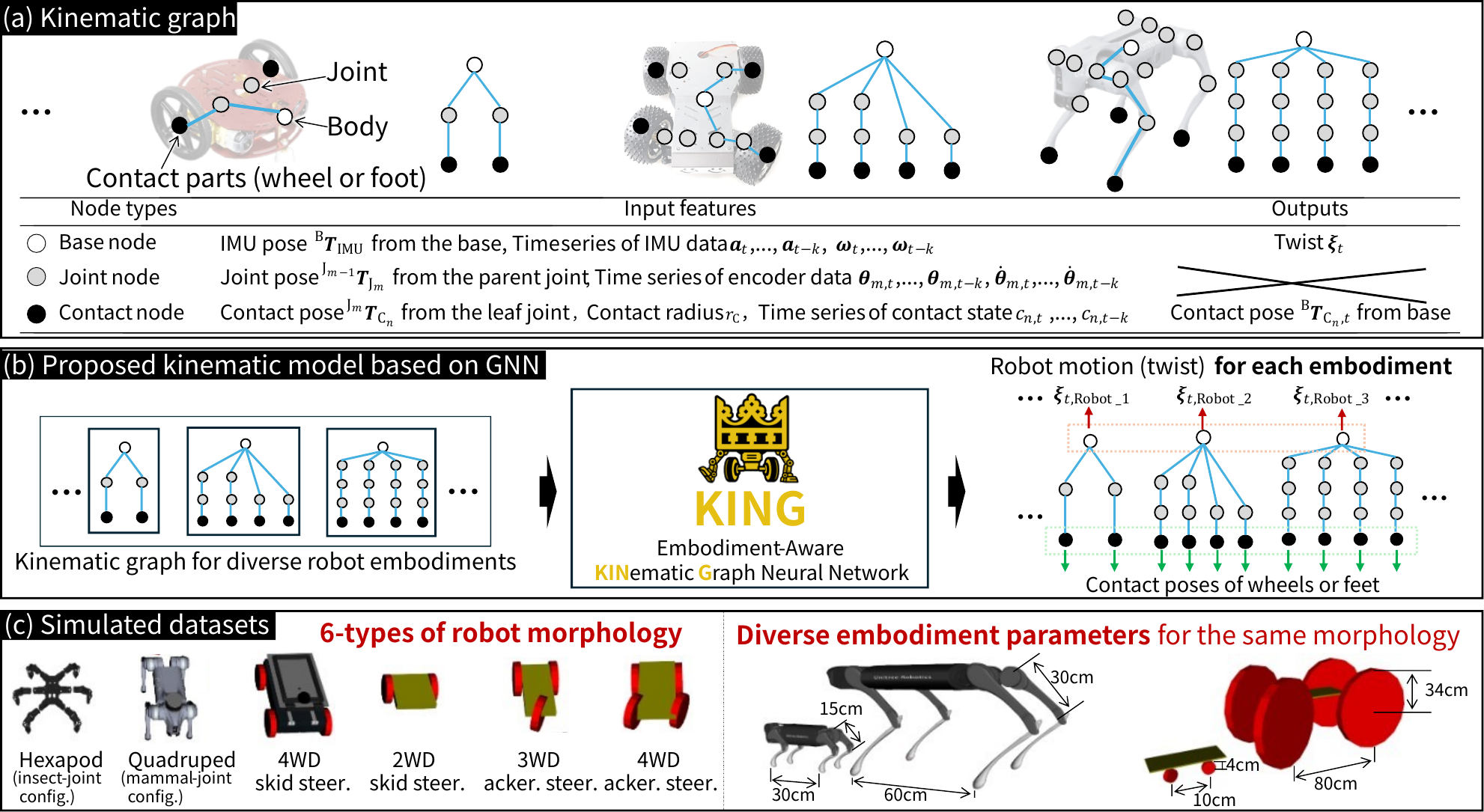}
  \caption{Overview of KING. We encode robot embodiments and local proprioceptive motion as unified kinematic graphs and train a GNN (KING) on various morphologies with diverse embodiment parameters to predict body twist and contact poses.}
  \label{fig:overview}
\end{figure*}

\subsection{Derivation of kinematic model equivalence between legs and wheels}\label{subsec:derivation_of_kinematic_model_equivalence}
Building on the unified kinematic representation using manipulator models in Sec.~\ref{subsec:unified_kinematic_representation}, we derive the kinematic model equivalence by showing that the wheel kinematic model can be obtained as a special case of a 1-DOF manipulator (1-DOF leg) kinematic model.

In the case of a 1-DOF manipulator with link length $l$ and joint angle $\theta$, its end-effector positions $x$ and $y$ are described as Eq.~\ref{eq:1DOF_arm}, as shown in Fig.~\ref{fig:1DOF_arm}(a):
\begin{align}
  \begin{bmatrix}
    x \\
    y
  \end{bmatrix}
  =
  \begin{bmatrix}
    l \cos\theta \\
    l \sin\theta
  \end{bmatrix}
  \label{eq:1DOF_arm}
\end{align}
We derive the end-effector velocities $\dot{x}$ and $\dot{y}$ (Eq.~\ref{eq:1DOF_v}) by taking the time derivative of Eq.~\ref{eq:1DOF_arm}:
\begin{align}
  \begin{bmatrix}
    \dot{x} \\
    \dot{y}
  \end{bmatrix}
  =
  \begin{bmatrix}
    -l\sin\theta \\
    ~~l \cos\theta
  \end{bmatrix} \dot{\theta}
  \label{eq:1DOF_v}
\end{align}
Substituting $\theta = 0$ (assuming ground contact) into Eq.~\ref{eq:1DOF_v}, $\dot{x}$ and $\dot{y}$ are described as Eq.~\ref{eq:1DOF_v_tokushu}:
\begin{align}
  \begin{bmatrix}
    \dot{x} \\
    \dot{y}
  \end{bmatrix}_{\theta = 0}
  =
  \begin{bmatrix}
    0 \\
    l \dot{\theta}
  \end{bmatrix} 
  \label{eq:1DOF_v_tokushu}
\end{align}
By replacing the link length $l$ with the wheel radius $r$, Eq.~\eqref{eq:1DOF_v_tokushu} becomes equivalent to the wheel kinematic model (i.e., $r\dot{\theta}$).
Therefore, the wheel kinematic model can be interpreted as that of a 1-DOF robotic arm (1-DOF leg) under the conditions $\theta = 0$, $l = r$, and continuous ground contact.
Furthermore, kinematic models of steered wheels are equivalent to kinematic models of 2-DOF robotic arms; however, the derivation is omitted for brevity.
% Therefore, we can interpret that the kinematic model of a wheel can be represented by a kinematic model of one DOF robotic arm (i.e., one DOF leg) in the conditions of $\theta = 0$, $l = r$, and continuous ground contact (Sec.~\ref{subsec:unified_kinematic_representation}).

\section{System overview}

The unified kinematic representation in Sec.~\ref{subsec:derivation_of_kinematic_model_equivalence} regards the ground-contact point of each wheel or foot as an end-effector in manipulator kinematics.
Motivated by this representation, our approach encodes kinematic structures and local proprioceptive motion signals from diverse wheeled and legged robots into a unified graph.
The number of nodes and edges in this graph depends on the robot morphology (e.g., the number of joints and ground-contact elements).
We therefore employ a graph neural network (GNN) to process variable-sized graphs and learn a versatile kinematic model representation, as message-passing-based GNNs naturally handle inputs with varying graph structures.

As shown in Fig.~\ref{fig:overview}(a), we construct a heterogeneous graph whose nodes correspond to the robot base, joints (actuators), and ground-contact elements (wheels or feet), referred to as \emph{Base}, \emph{Joint}, and \emph{Contact} nodes. Edges are connected based on the robot's kinematic connectivity. 
Each node is associated with features that encode (i) geometric and physical parameters (e.g., wheel radius and link length) and (ii) proprioceptive measurements (joint angles and angular velocities from encoders, linear acceleration and angular velocities from IMU). 
By composing local kinematic information along embodiment-specific connectivity, we designed the graph to provide a common representation that directly captures kinematic models of diverse robot embodiments. 
We assume that an IMU is mounted on the robot base.

\begin{figure*}[tb]
  \centering
  \includegraphics[width=1.0\linewidth]{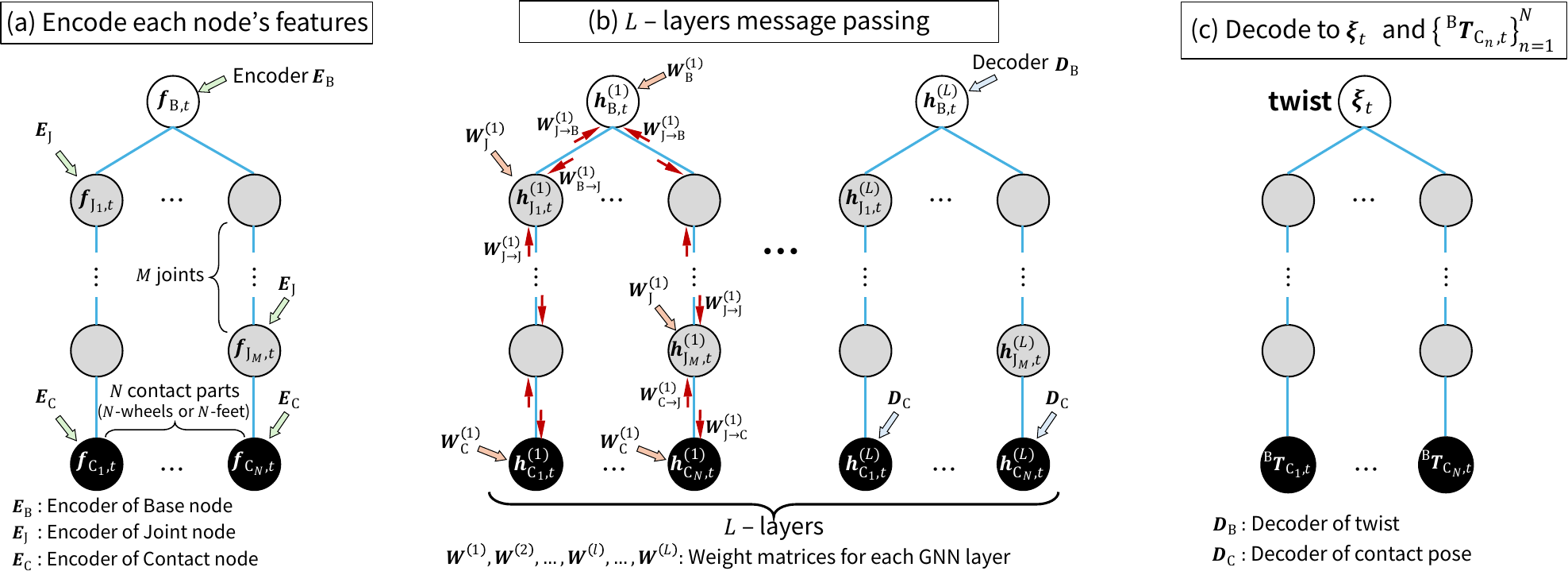}
  \caption{Overview of KING training via message-passing-based node regression.
           }
  \label{fig:gnn}
\end{figure*}

As illustrated in Fig.~\ref{fig:overview}(b), we jointly train a single kinematic model on data collected from diverse robot embodiments and apply the same model to previously unseen morphologies, thanks to the proposed unified kinematic representation and heterogeneous GNN architecture.
Because acquiring large-scale real-world motion data from many robot platforms is impractical, we leverage a physics-simulation environment to efficiently collect motion data across diverse robot embodiments, as shown in Fig.~\ref{fig:overview}(c).

\section{KING: Embodiment-Aware Kinematic Graph Neural Network}
\subsection{Graph structure with unified kinematic representation}\label{subsection:structure}
KING represents kinematic models of mobile robots with $N$ ground-contact elements (wheels or feet) and $M$ actuated joints as a graph (Fig.~\ref{fig:overview}(a)).
The graph consists of a single \emph{Base} node, $M$ \emph{Joint} nodes, and $N$ \emph{Contact} nodes, connected according to the robot's kinematic connectivity.
At each time step $t$, the GNN takes this graph as input and outputs (i) a body twist at the Base node, $\bm{\xi}_t \in \mathbb{R}^3$ (translational and rotational velocity in a tangent space), and (ii) the 6-DOF poses of the contact elements in the base frame, $\{^{\mathrm{B}}\bm{T}_{\mathrm{C}_{n},t}\}_{n=1}^{N}$ (Fig.~\ref{fig:overview}(b)).
The node features are designed to collectively provide the information necessary to represent wheel and leg kinematics as manipulator kinematics under the unified kinematic representation.

The Base node feature $\bm{f}_{\mathrm{B},t}$ consists of the 6-DOF relative pose $^{\mathrm{B}}{\bm T}_{\mathrm{IMU}}$ between the base frame and the IMU frame, the linear accelerations $\bm{a}_t \in \mathbb{R}^3$ and the angular velocities $\bm{\omega}_t \in \mathbb{R}^3$ measured by the IMU.

For the $m$-th Joint node, the feature $\bm{f}_{\mathrm{J}_m,t}$ consists of the initial 6-DOF relative pose $^{\mathrm{J}_{m-1}}\bm{T}_{\mathrm{J}_m}$ between adjacent joint frames (for $m=1$, $^{\mathrm{B}}\bm{T}_{\mathrm{J}_1}$ is set), together with the joint angle and angular velocity represented as 3D vectors, $\bm{\theta}_{m,t} \in \mathbb{R}^{3}$ and $\dot{\bm{\theta}}_{m,t} \in \mathbb{R}^{3}$.
This 3D vector explicitly encodes the rotation axis in the local joint frame to avoid aligning joint-axis directions (e.g., the DH convention).
For example, if a joint rotates about the $y$-axis, $\bm{\theta}_{m,t} = [0\ \theta_{m,t}\ 0]^\top$ and $\dot{\bm{\theta}}_{m,t} = [0\ \dot{\theta}_{m,t}\ 0]^\top$.
For wheel joints, we set $\bm{\theta}_{m,t} = \bm{0}$ based on the wheel--leg common formulation derived in Sec.~\ref{subsec:derivation_of_kinematic_model_equivalence}.

For the $n$-th Contact node, the feature $\bm{f}_{\mathrm{C}_n,t}$ consists of the 6-DOF relative pose $^{\mathrm{J}_{m}}{\bm T}_{\mathrm{C}_{n}}$ from the adjacent joint frame to the contact frame, the contact size $r_\mathrm{c}$, and a binary contact state $c_{t,n}$ (1 for contact, 0 for no contact).
Because foot shapes are often modeled as spheres in legged robot locomotion, we set $r_\mathrm{c}$ to the sphere radius for legged robots. For wheeled robots, we set the wheel radius as $r_\mathrm{c}$.
$r_\mathrm{c}$ encodes the local contact geometry required to represent the contact point velocity like an end-effector velocity in the case of manipulator models.
We assume that wheels are in continuous contact; thus, we set the contact state to $1$ for all wheels.
We designed the contact node to output the contact pose, which is the relative pose $^{\mathrm{B}}\bm{T}_{\mathrm{C}_{n},t}$ from the base frame to the contact frame, for learning embodiment-dependent geometric relationships in a shared output space. 

To improve robustness to sensor noise for sim-to-real deployment, time series data of all sensor measurements $\bm{a}_t$, $\bm{\omega}_t$, $\bm{\theta}_{m,t}$, $\dot{\bm{\theta}}_{m,t}$, and $c_{t,n}$ are used as each node feature by concatenating the current and past $k$ time steps. In our implementation, we set $k=2$; thus, the input feature dimensions are $\bm{f}_{\mathrm{B},t} \in \mathbb{R}^{24}$, $\bm{f}_{\mathrm{J}_m,t} \in \mathbb{R}^{24}$, and $\bm{f}_{\mathrm{C}_n,t} \in \mathbb{R}^{10}$.

\subsection{Training KING via Message-Passing Node Regression}
To train KING, we designed a message-passing-based loss function as a node-regression problem for the base and contact nodes.

\textbf{Message-passing procedure}
Given the input node features $\bm{f}_{\mathrm{B},t}$, $\{\bm{f}_{\mathrm{J}_m,t}\}_{m=1}^{M}$, and $\{\bm{f}_{\mathrm{C}_n,t}\}_{n=1}^{N}$ at time $t$, we first map them into latent embeddings using type-specific encoders with learnable weights
$\bm{E}_{\mathrm{B}}\in\mathbb{R}^{24\times 40}$, $\bm{E}_{\mathrm{J}}\in\mathbb{R}^{24\times 40}$, and $\bm{E}_{\mathrm{C}}\in\mathbb{R}^{10\times 40}$.
The encoder outputs are then projected into the 90-dimensional GNN hidden space using node-type-specific linear projection layers.
As shown in Fig.~\ref{fig:gnn}(a), (b), the initial node embeddings
$\bm{h}_{\mathrm{B},t}^{(1)}\in\mathbb{R}^{90}$,
$\{\bm{h}_{\mathrm{J}_m,t}^{(1)}\in\mathbb{R}^{90}\}_{m=1}^{M}$, and
$\{\bm{h}_{\mathrm{C}_n,t}^{(1)}\in\mathbb{R}^{90}\}_{n=1}^{N}$
are obtained by applying the respective encoders and projection layers
to the node features.

Message-passing is applied to propagate node features along the kinematic connectivity, as shown in Fig.~\ref{fig:gnn}(b).
Each message-passing weight matrix is
$\bm{W}^{(l)}\in\mathbb{R}^{90\times 90}$.
For example, the Base node embedding is updated by aggregating messages from adjacent Joint nodes:
\begin{align}
\bm{h}_{\mathrm{B},t}^{(l+1)}
&= \sigma \!\left(
\bm{W}_{\mathrm{B}}^{(l)} \bm{h}_{\mathrm{B},t}^{(l)}
\;+\;
\bigoplus_{m \in \mathcal{N}(\mathrm{B})}
\left(
\bm{W}_{\mathrm{J\!\to\!B}}^{(l)} \bm{h}_{\mathrm{J}_m,t}^{(l)}
\right)
\right),
\label{eq:king_message_passing_base_oplus}
\end{align}
where $\mathcal{N}(\mathrm{B})$ denotes the set of Joint nodes adjacent to the Base node,
$\sigma(\cdot)$ is an activation function (e.g., ReLU),
$\bigoplus$ is a permutation-invariant neighborhood aggregation operator (e.g., mean), and
$\bm{W}_{\mathrm{B}}^{(l)}$ and $\bm{W}_{\mathrm{J\!\to\!B}}^{(l)}$ are learnable weight matrices for transforming the Base node's own embedding and the messages from adjacent Joint nodes to the Base node, respectively.
Joint and Contact nodes are similarly updated by the same message-passing algorithm based on their local neighborhoods.
Because message-passing applies the same update rule regardless of the number of nodes/edges, the model naturally supports variable-DOF embodiments.

Figure~\ref{fig:gnn}(c) illustrates the decoding at the final GNN layer $L$.
We used learnable linear decoders $\bm{D}_{\mathrm{B}}\in\mathbb{R}^{90\times 3}$ and $\bm{D}_{\mathrm{C}}\in\mathbb{R}^{90\times 6}$ to decode $\bm{h}_{\mathrm{B},t}^{(L)}$ and $\{\bm{h}_{\mathrm{C}_n,t}^{(L)}\}_{n=1}^{N}$ to the base twist $\bm{\xi}_t$ and the contact poses $\{^{\mathrm{B}}\bm{T}_{\mathrm{C}_{n},t}\}_{n=1}^{N}$, respectively.

\textbf{Loss function:}
Based on the above message passing procedure, we define the loss function $\mathcal{L}$ as a sum of the weighted mean squared errors (MSEs) for the base twist loss $\mathcal{L}_{\xi}$ and the contact pose loss $\mathcal{L}_{\mathrm{c}}$:
\begin{align}
\mathcal{L}
  &= \mathcal{L}_{\xi} + \mathcal{L}_{\mathrm{c}},
  \label{eq:total_loss}
\\
\mathcal{L}_{\xi}
  &= \frac{1}{S}\sum_{t=1}^{S}
     \Bigl(
       w^{\mathrm{trans}}_{\xi}\,e^{\mathrm{trans}}_{t,\xi}
       \;+\;
       w^{\mathrm{rot}}_{\xi}\,e^{\mathrm{rot}}_{t,\xi}
       \;+\;
       \rho(w^{\mathrm{trans}}_{\xi}, w^{\mathrm{rot}}_{\xi})
     \Bigr).
  \label{eq:twist_loss_uncertainty}
\\
\mathcal{L}_{\mathrm{c}}
  &= \frac{1}{S}\sum_{t=1}^{S}
     \Big(
       w^{\mathrm{trans}}_{\mathrm{c}}\,e^{\mathrm{trans}}_{t,\mathrm{c}} + w^{\mathrm{rot}}_{\mathrm{c}} \, e^{\mathrm{rot}}_{t,\mathrm{c}}
     \Big).
  \label{eq:contact_loss_constant}
\end{align}
where $e^{\mathrm{trans}}_{t,\xi}$ and $e^{\mathrm{rot}}_{t,\xi}$ denote the squared errors of the translational and rotational components of $\bm{\xi}_t$, respectively.
$w^{\mathrm{trans}}_{\xi}$ and $w^{\mathrm{rot}}_{\xi}$ are the corresponding weight parameters to balance the translational and rotational scales.
In particular, we adopt the uncertainty-based weighting method~\cite{kendall2018multi} for the base twist loss, where $\rho(w^{\mathrm{trans}}_{\xi}, w^{\mathrm{rot}}_{\xi})$ is a regularization term that encourages $w^{\mathrm{trans}}_{\xi}$ and $w^{\mathrm{rot}}_{\xi}$ to prevent trivial solutions.
$e^{\mathrm{trans}}_{t,\mathrm{c}}$ and $e^{\mathrm{rot}}_{t,\mathrm{c}}$ denote the squared errors of the translational and rotational components of the contact pose outputs, respectively.
$w^{\mathrm{trans}}_{\mathrm{c}}$ (e.g., 0.7) and $w^{\mathrm{rot}}_{\mathrm{c}}$ (e.g., 0.001) are set to the constant values to balance the scales of the contact pose loss components.
$S$ is the number of training samples.

We use AdamW to minimize \EQ{eq:total_loss} with learning rate $10^{-4}$, batch size 128, and 40 epochs.

\subsection{Automatic data collection for diverse robot embodiments}\label{subsection:dataset}
Collecting large-scale real-world motion data for many robot embodiments is expensive and time-consuming.
Following recent simulation-based data generation approaches~\cite{wassermanlegolas,xiao2025anycar}, we build an automatic data collection pipeline in a physics simulator (Gazebo) to efficiently collect a large training dataset across diverse embodiments.
Specifically, we use six different robot morphologies: 2WD and 4WD skid-steering robots, 3WD and 4WD Ackermann-steering robots, quadruped and hexapod robots, as shown in Fig.~\ref{fig:overview}(c).
Note that the 2WD and 4WD skid-steering robots are driven by independently actuated left and right wheel groups, while the 3WD and 4WD Ackermann-steering robots have steering mechanisms on their front wheels.
The joint configurations of the quadruped and hexapod robots differ in addition to the number of legs: the quadruped has a mammal-like leg with three joints (i.e., hip, thigh, and knee), while the hexapod has an insect-like leg with three joints (i.e., coxa, femur, and tibia).
Furthermore, we conducted domain randomization of the embodiment parameters for all these robots to increase the diversity of the training data; thus, we can more easily collect various embodiments of the six robot morphologies.
For example, as illustrated in Fig.~\ref{fig:overview}(c), the randomized quadruped embodiments include robots with approximately two-fold differences in link lengths, and the randomized 4WD Ackermann-steering embodiments include robots with wheel diameters differing by more than eight-fold.
To collect diverse motion data, each robot with different embodiments is controlled to execute various planar motion trajectories, including straight, turning, and slalom-like motions, with acceleration and deceleration.
Furthermore, data augmentation is applied to add noise to all proprioceptive measurements and embodiment parameters, thereby increasing the diversity of the training data and improving the model's robustness for the sim-to-real gap.

\begin{table}[tb]
  \caption{Cross-class transfer results. In both cases, increasing source-class data ratio reduces target-class errors, indicating positive transfer between wheeled and legged data.}
  \label{tab:cross_class_transfer}
  \centering
  \scriptsize
  \setlength{\tabcolsep}{3.0pt}
  \renewcommand{\arraystretch}{1.08}

  \begin{threeparttable}
    \resizebox{\columnwidth}{!}{%
      \begin{tabular}{c|cc|cc}
        \toprule
        Src. data ratio [\%] &
        \multicolumn{2}{c|}{train: wheel, test: leg} &
        \multicolumn{2}{c}{train: leg, test: wheel} \\
        \cline{2-5}
        & $\mathrm{RMSE}_{\xi,\mathrm{t}}$ & $\mathrm{RMSE}_{\xi,\mathrm{r}}$
        & $\mathrm{RMSE}_{\xi,\mathrm{t}}$ & $\mathrm{RMSE}_{\xi,\mathrm{r}}$ \\
        \midrule
        0   & 0.206 m/s & 0.184 rad/s & 0.174 m/s & 0.133 rad/s \\
        10  & 0.168 m/s & 0.070 rad/s & 0.109 m/s & 0.031 rad/s \\
        20  & 0.160 m/s & 0.051 rad/s & 0.107 m/s & 0.030 rad/s \\
        \bottomrule
      \end{tabular}%
    }

    \begin{tablenotes}[flushleft]
      \footnotesize
      \item[1] $\mathrm{RMSE}_{\xi,\mathrm{t}}$ and $\mathrm{RMSE}_{\xi,\mathrm{r}}$ are the translational [m/s] and rotational [rad/s] twist RMSEs, respectively.
    \end{tablenotes}
  \end{threeparttable}
\end{table}
%  ($\mathrm{RMSE}_{\xi,\mathrm{t}}[\mathrm{m/s}]$, $\mathrm{RMSE}_{\xi,\mathrm{r}}[\mathrm{rad/s}]$)
\begin{table*}[t]
  \centering
  \caption{Cross-validation results across robot morphologies in the simulation environment.}
  \label{tab:ablation_study_of_identicalMLP3}
  \begin{threeparttable}
    \setlength{\tabcolsep}{0.74mm}
    \begin{tabular}{cc|cc|cc|cc|cc|cc|cc|}
      \hline
      \multicolumn{2}{c|}{} &
      \multicolumn{2}{c|}{2WD skid-steer} &
      \multicolumn{2}{c|}{4WD skid-steer} &
      \multicolumn{2}{c|}{3WD Ackermann} &
      \multicolumn{2}{c|}{4WD Ackermann} &
      \multicolumn{2}{c|}{Quadruped} &
      \multicolumn{2}{c|}{Hexapod} \\
      \cline{3-14}
      \multicolumn{2}{c|}{Method~/~Morphology} &
      $\mathrm{RMSE}_{\xi,\mathrm{t}}$ & \cellcolor[HTML]{EFEFEF}$\mathrm{RMSE}_{\xi,\mathrm{r}}$ &
      $\mathrm{RMSE}_{\xi,\mathrm{t}}$ & \cellcolor[HTML]{EFEFEF}$\mathrm{RMSE}_{\xi,\mathrm{r}}$ &
      $\mathrm{RMSE}_{\xi,\mathrm{t}}$ & \cellcolor[HTML]{EFEFEF}$\mathrm{RMSE}_{\xi,\mathrm{r}}$ &
      $\mathrm{RMSE}_{\xi,\mathrm{t}}$ & \cellcolor[HTML]{EFEFEF}$\mathrm{RMSE}_{\xi,\mathrm{r}}$ &
      $\mathrm{RMSE}_{\xi,\mathrm{t}}$ & \cellcolor[HTML]{EFEFEF}$\mathrm{RMSE}_{\xi,\mathrm{r}}$ &
      $\mathrm{RMSE}_{\xi,\mathrm{t}}$ & \cellcolor[HTML]{EFEFEF}$\mathrm{RMSE}_{\xi,\mathrm{r}}$ \\
      \hline \hline

      \multicolumn{2}{c|}{$K=0\%$~(zero-shot)} &
      0.074 & \cellcolor[HTML]{EFEFEF}{0.007} &
      0.108 & \cellcolor[HTML]{EFEFEF}{0.005} &
      0.060 & \cellcolor[HTML]{EFEFEF}{0.004} &
      0.072 & \cellcolor[HTML]{EFEFEF}{0.004} &
      0.210 & \cellcolor[HTML]{EFEFEF}{0.011} &
      0.130 & \cellcolor[HTML]{EFEFEF}{0.008} \\

      \multicolumn{2}{c|}{$K=0.1\%$~(FT with 1~min data)} &
      0.038 & \cellcolor[HTML]{EFEFEF}{0.005} &
      0.031 & \cellcolor[HTML]{EFEFEF}{0.004} &
      0.030 & \cellcolor[HTML]{EFEFEF}{0.003} &
      0.025 & \cellcolor[HTML]{EFEFEF}{0.003} &
      0.142 & \cellcolor[HTML]{EFEFEF}{0.011} &
      0.070 & \cellcolor[HTML]{EFEFEF}{0.004} \\

      \multicolumn{2}{c|}{$K=0.5\%$~(FT with 5~min data)} &
      0.029 & \cellcolor[HTML]{EFEFEF}{0.004} &
      0.023 & \cellcolor[HTML]{EFEFEF}{0.003} &
      0.020 & \cellcolor[HTML]{EFEFEF}{0.003} &
      0.019 & \cellcolor[HTML]{EFEFEF}{0.003} &
      0.082 & \cellcolor[HTML]{EFEFEF}{0.009} &
      0.049 & \cellcolor[HTML]{EFEFEF}{0.004} \\

      \multicolumn{2}{c|}{$K=1\%$~(FT with 10~min data)} &
      0.024 & \cellcolor[HTML]{EFEFEF}{0.008} &
      0.016 & \cellcolor[HTML]{EFEFEF}{0.004} &
      0.016 & \cellcolor[HTML]{EFEFEF}{0.003} &
      0.016 & \cellcolor[HTML]{EFEFEF}{0.002} &
      0.067 & \cellcolor[HTML]{EFEFEF}{0.008} &
      0.038 & \cellcolor[HTML]{EFEFEF}{0.003} \\

      \multicolumn{2}{c|}{Using all morphologies} &
      0.022 & \cellcolor[HTML]{EFEFEF}{0.008} &
      0.023 & \cellcolor[HTML]{EFEFEF}{0.020} &
      0.034 & \cellcolor[HTML]{EFEFEF}{0.020} &
      0.051 & \cellcolor[HTML]{EFEFEF}{0.027} &
      0.030 & \cellcolor[HTML]{EFEFEF}{0.021} &
      0.038 & \cellcolor[HTML]{EFEFEF}{0.018} \\
      \hline
    \end{tabular}

  \end{threeparttable}
\end{table*}

\section{Experimental results}
In this section, we evaluate KING from two perspectives.
First, we empirically validate the unified wheel--leg kinematic formulation derived in Sec.~\ref{section:UKMR} by testing whether training data from wheeled robots improves body-twist estimation for legged robots, and vice versa.
Second, we assess KING's generalization to unseen robot morphologies via cross-validation.

\subsection{Validation of the unified kinematic model representation}
\label{subsec:exp_wheel_leg_equivalence}
We demonstrate the validity of the unified kinematic model representation for legged and wheeled robots described in Sec.~\ref{section:UKMR}. 
If both kinematic models are described in the common representation, motion data collected from one class (wheeled or legged robots) should help train KING to improve estimation accuracy for the other class. 
Conversely, if they are not related, additional data from the other class could degrade model accuracy due to mismatched kinematic representations.
To validate this hypothesis, we train KING on data from a single class and evaluate it on the other, while progressively increasing the amount of training data.

We used the dataset (Sec.~\ref{subsection:dataset}) and split it into a wheeled class (2WD and 4WD skid-steering robots, and 3WD and 4WD ackermann robots) and a legged class (quadruped and hexapod robots).
We then performed two cross-class transfer experiments: (i) train KING primarily on the wheeled class and evaluate on the legged class, and (ii) train KING primarily on the legged class and evaluate on the wheeled class.
In each experiment, we progressively increase the amount of training data from the source class ($0\%$, $10\%$, and $20\%$) while keeping the target-class dataset fixed.
To ensure that the model observes the contact-state patterns of both wheeled and legged robots, we include a tiny fraction ($0.01\%$) of the target-class data in all settings.
The $0\%$ source-class setting therefore corresponds to training only on this tiny target-class subset, confirming that such negligible target data alone is insufficient to learn the target-class kinematics.

Table~\ref{tab:cross_class_transfer} summarizes the cross-class transfer results regarding the RMSE of translational and rotational twist on the target dataset.
In both legged and wheeled classes, increasing the amount of source-class training data (leg or wheel) reduces the target-set (wheel or leg) RMSEs.
In the $0\%$ source-class setting, the resulting errors are substantially larger than those in the $10\%$ and $20\%$ source-class settings, showing that the tiny amount of target-class data alone is insufficient to learn the target-class kinematics; thus, we consider that setting the tiny fraction of the target-class data to $0.01\%$ was a reasonable choice.
These results are consistent with our hypothesis, adding more source-class data (wheeled or legged) systematically improves estimation accuracy on the other class, rather than degrading it. This indicates that both kinematic models are compatible under the proposed common representation, providing empirical support for the unified kinematic model representation derived in Sec.~\ref{section:UKMR}.

\subsection{Cross-validation on unseen robot morphologies}
\label{subsec:cross_validation}
We evaluate how well KING generalizes to robot morphologies not seen during pretraining in zero-shot and few-shot adaptation settings.
We used the six morphology datasets described in Fig.~\ref{fig:overview}(c).
We performed leave-one-morphology-out cross-validation: in each fold, one morphology is designated as the \emph{target} morphology and excluded from the multi-morphology training data, while the remaining five morphologies are used to obtain a pretrained model.
This procedure is repeated for each target morphology, and performance is evaluated on the corresponding target datasets.
After pretraining, we fine-tune the model using a small adaptation dataset from the target morphology and evaluate on a disjoint target-morphology test split.
We vary the size of the adaptation set (denoted by $K~[\%]$) to show how much data is needed to fine-tune the model on unseen robot morphologies.
As a complementary reference, we also train an all-morphology model on $20\%$ of the full dataset.
This setting is not part of the leave-one-morphology-out protocol because all morphologies are included during training.
Instead, it evaluates within-morphology generalization to unseen embodiment parameters, since the validation split contains robots with the same morphologies but randomized embodiment parameters (e.g., wheel radii, link lengths, and IMU placement) that differ from those used for training.

Table~\ref{tab:ablation_study_of_identicalMLP3} summarizes the results in terms of twist RMSE.
The \textit{using all morphologies} row provides a reference performance when the target morphology is already included in the training dataset; however, the embodiment parameters are unseen.
Its low RMSEs across morphologies indicate that KING can generalize to variations in embodiment parameters within known morphologies.
Even with $K=0.1\%$ (approximately 1 minute of target data), few-shot fine-tuning yields a substantial performance gain across all morphologies, and mostly matches the all-morphology baseline for all targets except the quadruped.
With $K\ge 0.5\%$ (5 minutes or more), the improvement becomes more gradual, and the quadruped performance also approaches saturation, indicating that effective adaptation can be achieved with only a small amount of target data.
In the zero-shot setting ($K=0\%$), the 2WD skid-steering and Ackermann-drive robots achieve lower errors, likely because their motions involve fewer nonlinear effects, such as severe slip, than the more dynamically complex morphologies.
Overall, these results show that KING can rapidly adapt to unseen robot morphologies with minimal target-domain data.

\begin{figure*}[tb]
  \centering
  \includegraphics[width=1.0\linewidth]{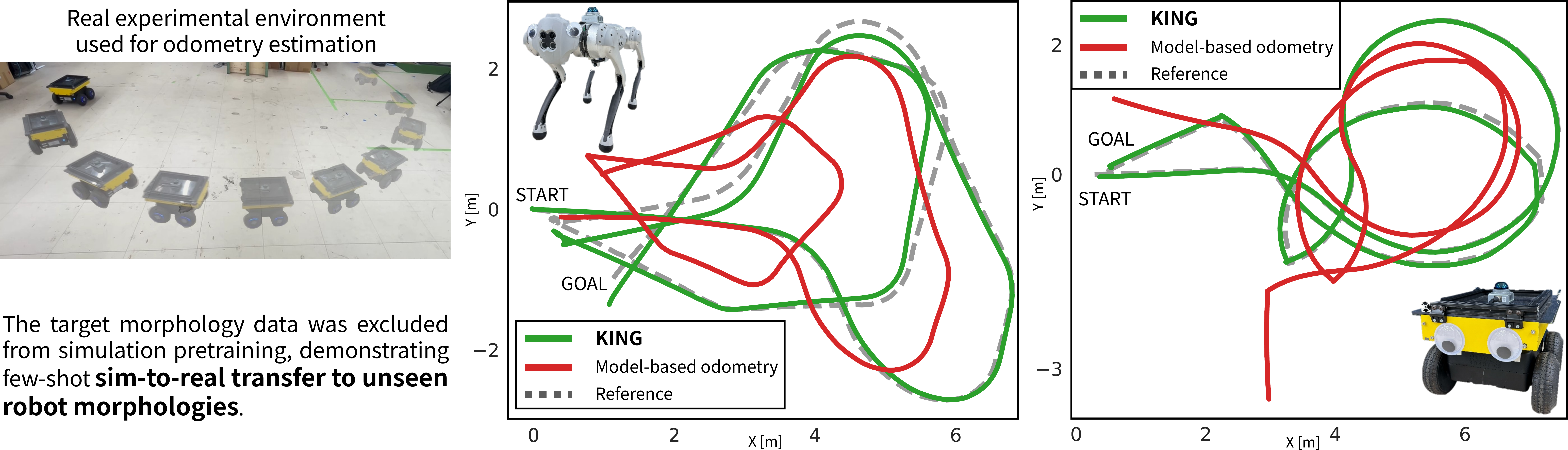}
  \caption{Real experimental environment, estimated trajectories on robots whose morphologies are unseen during pretraining.}
  \label{fig:jikki_traj}
\end{figure*}

\subsection{Sim-to-Real transfer to unseen robot morphologies}
\label{subsec:real_cross_validation}
Extending Sec.~\ref{subsec:cross_validation} to sim-to-real transfer, we evaluated few-shot adaptation on unseen real robots.
For each testbed, we pretrained the model on simulation data that excluded its morphology, then fine-tuned this model using only a small amount of real data to mitigate the sim-to-real gap (e.g., friction parameters).
We used two testbeds: a four-wheeled skid-steering robot (Rover Robotics Rover Mini) and a quadruped robot (Unitree Go1).
For both platforms, we collected approximately 1-min motion datasets by driving these robots on a flat indoor floor (Fig.~\ref{fig:jikki_traj}) and fine-tuned the pretrained model on these data.
According to the simulation results in Sec.~\ref{subsec:cross_validation}, we expected that a small amount of 1-minute data is sufficient to achieve fine-tuning that significantly improves estimation accuracy on unseen real robots.
To obtain reference data for fine-tuning, we ran LiDAR--IMU SLAM using an omni-directional LiDAR (Livox MID-360) and computed a reference twist from the resulting SLAM trajectory.
We used the IMU embedded in the MID-360 and joint/wheel encoder values from the robots as input data for KING.
As a model-based baseline, we used an analytic kinematic model in which translational velocity is computed from kinematics (Eq.~\ref{eq:leg_odom} or Eq.~\ref{eq:wheel_odom}), and rotational velocity is provided by the IMU gyroscope.
This baseline is sensitive to terrain interactions such as slip, which are difficult to capture with a purely analytic model.

% \paragraph{Results.}
Table~\ref{tab:jikki_twist_rmse} summarizes twist RMSE on both robots. 
For the 4WD skid-steering robot, while real-world translational and rotational RMSEs are $0.038 \mathrm{m/s}$ and $0.022 \mathrm{rad/s}$, the simulation results for the same morphology in Table~\ref{tab:ablation_study_of_identicalMLP3} (Using all morphologies case) show RMSEs of $0.034 \mathrm{m/s}$ and $0.027 \mathrm{rad/s}$. For the quadruped robot, the real-world RMSEs are $0.049 \mathrm{m/s}$ and $0.026 \mathrm{rad/s}$, while the simulation results for the same morphology show RMSEs of $0.030 \mathrm{m/s}$ and $0.021 \mathrm{rad/s}$.
Therefore, this result shows that the proposed model with few-shot adaptation achieves accuracy comparable to that of the simulation results.

Fig.~\ref{fig:jikki_traj} compares the trajectories estimated by \textit{KING} and \textit{Model-based odometry} with the reference trajectories.
Model-based odometry exhibits noticeable drift, likely due to slip and other contact effects, whereas KING remains closer to the reference trajectories, indicating that the learned model better captures these interactions.
These results show that a brief 1-minute dataset is sufficient for effective few-shot sim-to-real adaptation of KING and for accurate odometry estimation, consistent with the simulation results.
This indicates that a kinematic model trained primarily on simulation can be transferred to real robots with morphologies unseen during pretraining through minimal real-world fine-tuning.

\begin{table}[t]
  \centering
  \caption{Twist RMSE on unseen real robots.}
  \label{tab:jikki_twist_rmse}
  \scriptsize
  \setlength{\tabcolsep}{3pt}
  \renewcommand{\arraystretch}{1.05}
  \resizebox{\columnwidth}{!}{%
    \begin{tabular}{l|cc|cc}
      \hline
      & \multicolumn{2}{c|}{4WD skid-steer.} & \multicolumn{2}{c}{Quadruped} \\
      \cline{2-5}
      & $\mathrm{RMSE}_{\xi,\mathrm{t}}$ & $\mathrm{RMSE}_{\xi,\mathrm{r}}$
      & $\mathrm{RMSE}_{\xi,\mathrm{t}}$ & $\mathrm{RMSE}_{\xi,\mathrm{r}}$ \\
      \hline \hline
      KING (FT with 1-min data) &
      0.038 m/s & 0.022 rad/s &
      0.049 m/s & 0.026 rad/s \\
      \hline
    \end{tabular}%
  }
\end{table}

\section{CONCLUSIONS}
We presented KING, an embodiment-aware kinematic graph neural network for learning kinematic models applicable to both legged and wheeled robots.
We first established a unified kinematic model, showing that the kinematics of wheels and legs can be represented within a common framework based on manipulator kinematics.
Based on this unified representation, we encoded robot embodiments and local proprioceptive motion signals in a common graph and trained KING on large-scale simulated data spanning diverse embodiments.
Through cross-validation, we demonstrated that KING can be efficiently adapted to morphologies unseen during pretraining, requiring only about one minute of target data for effective fine-tuning.
On real robots, this few-shot adaptation enabled accurate odometry estimation in real environments and outperformed the model-based baseline.

Future work will extend KING with online learning, building on prior approaches~\cite{okawara2024neuralwheel,okawara2025tightlyleg}, toward an odometry foundation model capable of adapting to heterogeneous robot embodiments and varying terrain conditions.

\addtolength{\textheight}{-12cm}   % This command serves to balance the column lengths

%%%%%%%%%%%%%%%%%%%%%%%%%%%%%%%%%%%%%%%%%%%%%%%%%%%%%%%%%%%%%%%%%%%%%%%%%%%%%%%%

%%%%%%%%%%%%%%%%%%%%%%%%%%%%%%%%%%%%%%%%%%%%%%%%%%%%%%%%%%%%%%%%%%%%%%%%%%%%%%%%



% \section*{ACKNOWLEDGMENT}

% The preferred spelling of the word acknowledgment in America is without an e after the g. Avoid the stilted expression, One of us (R. B. G.) thanks . . .  Instead, try R. B. G. thanks. Put sponsor acknowledgments in the unnumbered footnote on the first page.

%%%%%%%%%%%%%%%%%%%%%%%%%%%%%%%%%%%%%%%%%%%%%%%%%%%%%%%%%%%%%%%%%%%%%%%%%%%%%%%%

% References are important to the reader; therefore, each citation must be complete and correct. If at all possible, references should be commonly available publications.

\balance

% --- BibTeX で参考文献を出力する設定 ---
% .bib ファイル（例: references.bib）を用意し、本文では \cite{<key>} で引用します。
% コンパイル手順（BibTeX）: platex -> (up)bibtex -> platex -> platex
%
% この zip には references.bib を同梱しています。
% 自分の .bib を使う場合は次行のファイル名を置き換えてください（拡張子 .bib は不要）。
\bibliographystyle{IEEEtran}
% 使う .bib ファイル名（拡張子 .bib は不要）
% ※この zip には references.bib を同梱しています。
\bibliography{references}

\end{document}